%% file: main.tex
\documentclass{article}

\usepackage{microtype}
\usepackage{graphicx}
\usepackage{booktabs} 
\usepackage{algorithmic,algorithm}
\usepackage{amsfonts}
\usepackage{amsmath}
\usepackage{amsopn}

\usepackage{amssymb}
\usepackage{amsthm}
\usepackage{bm}
\usepackage{booktabs}
\usepackage{color}
\usepackage{enumitem}
\usepackage{float}
\usepackage{graphicx}
\usepackage{hyperref}
\usepackage{mathrsfs}
\usepackage{mathtools}
\usepackage{multirow}
\usepackage{siunitx}
\usepackage{stfloats}
\usepackage{thmtools,thm-restate}
\usepackage{threeparttable}
\usepackage{verbatim}
\usepackage{wrapfig}
\usepackage{url}
\usepackage{bbding}
\usepackage{diagbox}
\usepackage{stackengine}

\usepackage{subcaption}
\usepackage{graphicx}
\usepackage{float}
\usepackage{makecell}
\usepackage{xcolor}         
\usepackage[most]{tcolorbox} 

\usepackage{listings}
\definecolor{keywordcolor}{rgb}{0.0, 0.1, 0.6}   
\definecolor{tacticcolor}{rgb}{0.0, 0.1, 0.6}    
\definecolor{commentcolor}{rgb}{0.1, 0.5, 0.1}   
\definecolor{symbolcolor}{rgb}{0.0, 0.1, 0.6}    
\definecolor{sortcolor}{rgb}{0.0, 0.1, 0.6}      
\definecolor{attributecolor}{rgb}{0.7, 0.1, 0.1} 

\lstnewenvironment{leancode}{\lstset{language=lean}}{}
\usepackage{hyperref}

\newcommand\xrowht[2][0]{\addstackgap[.5\dimexpr#2\relax]{\vphantom{#1}}}

\usepackage[accepted]{icml2025}

\usepackage{amsmath}
\usepackage{amssymb}
\usepackage{mathtools}
\usepackage{amsthm}

\usepackage[capitalize,noabbrev]{cleveref}

\theoremstyle{plain}

\theoremstyle{definition}

\theoremstyle{remark}

\usepackage[textsize=tiny]{todonotes}

\icmltitlerunning{Learning an Effective Premise Retrieval Model for Efficient Mathematical Formalization}

\begin{document}

\twocolumn[
\icmltitle{Learning an Effective Premise Retrieval Model for Efficient\\ Mathematical Formalization}



\icmlsetsymbol{equal}{*}

\begin{icmlauthorlist}
\icmlauthor{Yicheng Tao}{equal,ai,math}
\icmlauthor{Haotian Liu}{equal,ai}
\icmlauthor{Shanwen Wang}{math}
\icmlauthor{Hongteng Xu}{ai,lab,moe}
\end{icmlauthorlist}

\icmlaffiliation{ai}{Gaoling School of Artificial Intelligence, Renmin University of China}
\icmlaffiliation{math}{School of Mathematics, Renmin University of China}
\icmlaffiliation{lab}{Beijing Key Laboratory of Research on Large Models and Intelligent Governance}
\icmlaffiliation{moe}{Engineering Research Center of Next-Generation Intelligent Search and Recommendation, MOE}

\icmlcorrespondingauthor{Shanwen Wang}{s\_wang@ruc.edu.cn}
\icmlcorrespondingauthor{Hongteng Xu}{hongtengxu@ruc.edu.cn}

\icmlkeywords{Machine Learning, ICML}

\vskip 0.3in
]



\printAffiliationsAndNotice{\icmlEqualContribution} 

\begin{abstract}
Formalized mathematics has recently garnered significant attention for its ability to assist mathematicians across various fields.
Premise retrieval, as a common step in mathematical formalization, has been a challenge, particularly for inexperienced users.
Existing retrieval methods that facilitate natural language queries require a certain level of mathematical expertise from users, while approaches based on formal languages (e.g., Lean) typically struggle with the scarcity of training data, hindering the training of effective and generalizable retrieval models.
In this work, we introduce a novel method that leverages data extracted from Mathlib to train a lightweight and effective premise retrieval model.
In particular, the proposed model embeds queries (i.e., proof state provided by Lean) and premises in a latent space, featuring a tokenizer specifically trained on formal corpora.
The model is learned in a contrastive learning framework, in which a fine-grained similarity calculation method and a re-ranking module are applied to enhance the retrieval performance.
Experimental results demonstrate that our model outperforms existing baselines, achieving higher accuracy while maintaining a lower computational load.
We have released an open-source search engine based on our retrieval model at \url{https://premise-search.com/}.
The source code and the trained model can be found at \url{https://github.com/ruc-ai4math/Premise-Retrieval}.
\end{abstract}

\input{sections/introduction}

\input{sections/preliminaries}

\input{sections/relatedwork}
\input{sections/method}

\input{sections/experiments}

\input{sections/conclusion.tex}



\section*{Acknowledgements}
Yicheng Tao and Shanwen Wang are supported by National Key Technologies R\&D Program of China, Grant No. 2024YFA1014001.
Haotian Liu and Hongteng Xu are supported by Beijing Natural Science Foundation (Grant No. L233008), the fund for building world-class universities of Renmin University of China, and the funds from Beijing Key Laboratory of Research on Large Models and Intelligent Governance, and Engineering Research Center of Next-Generation Intelligent Search and Recommendation, Ministry of Education.
We thank Beijing International Center for Mathematical Research(BICMR) at Peking University and Institute for Mathematical Sciences(IMS) at National University of Singapore for organizing Lean workshops, which provides great opportunities for learning and communication.

\section*{Impact Statement}
The formalization community has been exploring the potential of AI tools as assistants. In this work we develop an intelligent search engine for formal theorems based on proof states, aiming to accelerating the process of mathematical formalization.
\bibliography{references}
\bibliographystyle{icml2025}

\newpage
\appendix
\onecolumn


\input{sections/appendix}

\end{document}

%% file: sections/introduction.tex
\section{Introduction}
Formalized mathematics has recently attracted significant attention. 
It helps verify existing mathematical results, identify errors in the literature, and has the potential to accelerate the peer-review process for mathematical papers. 
This process involves proving natural language theorems within a strict formal logical framework. 
Interactive theorem provers (ITPs), or proof assistants, are commonly used for this purpose.
In recent years, mathematicians have actively participated in developing large formalized theorem libraries using ITPs such as Lean~\cite{lean_ori_de2015lean}, Coq~\cite{coq_barras1997coq}, and Isabelle~\cite{isabelle_nipkow2002isabelle}.

\begin{figure}[t]
    \centering
    \includegraphics[width=0.9\linewidth]{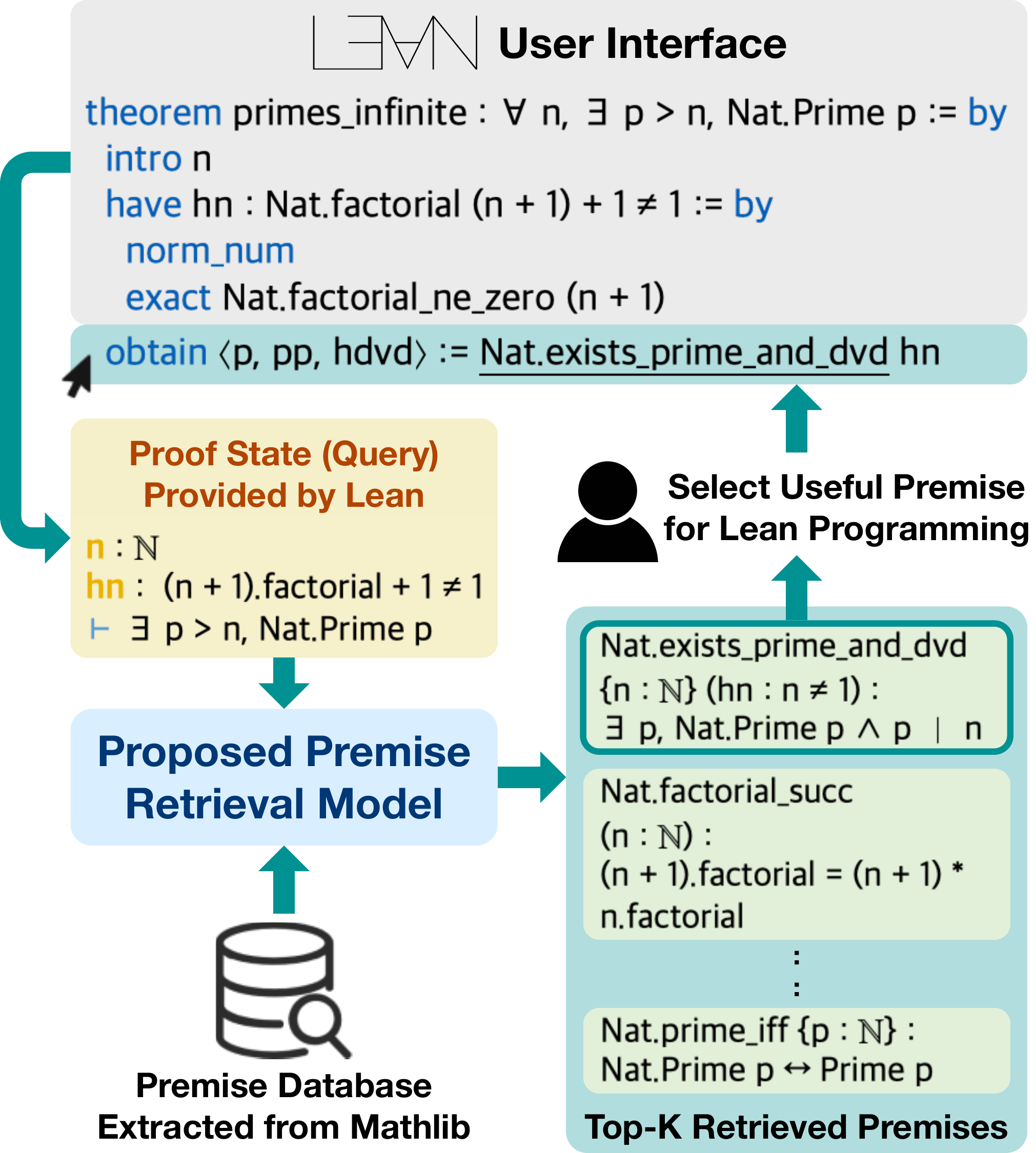}
    \caption{A schematic diagram illustrating user interactions with an ITP (e.g., Lean) given our premise retrieval model.}
    \label{fig:display}
\end{figure}

However, the process of mathematics formalization often demands extensive experience with formalization and a high level of mathematical expertise.
Therefore, there has been a sustained demand for automated auxiliary tools in the formalization field, such as premise retrievers. 
As illustrated in Figure~\ref{fig:display}, premise retrieval aims to retrieve useful premises based on the current proof state provided by the formalization platform.
An effective premise retrieval system can help mathematicians quickly identify relevant theorems, thereby accelerating the proof process. 
For example, LeanDojo~\cite{lean_dojo_yang2024leandojo} shows that a retrieval-augmented generator achieves a higher proof pass rate compared to a standalone generator without premise selection.

The early premise retrieval methods leverage internal APIs of ITPs to identify applicable theorems through pattern-matching algorithms.
These methods, however, rely on strict matching criteria and may miss relevant theorems.
Recently, learning-based methods~\cite{lean_dojo_yang2024leandojo, hammer_mikula2023magnushammer} have been proposed to encode formalized theorems in a latent space and retrieve them by computing their embedding similarities. 
The learning-based methods can be further categorized based on the query language: natural language queries or formal proof state queries. 
Natural language-based retrieval~\cite{pku-gao-etal-2024-semantic-search} requires users to have a sufficient mathematical understanding of the current proof state, posing a significant challenge for beginners. 
In contrast, formal language-based retrieval is more user-friendly, but it often relies on fine-tuning pre-trained natural language models~\cite{lean_dojo_yang2024leandojo} because of the scarcity of formal proof data. 
As a result, the fundamental differences between formal and natural languages introduce a semantic gap, leading to suboptimal performance. 

The above challenges motivate us to design and learn a more effective premise retrieval system that directly leverages the formal proof state to retrieve premises, improving accuracy and usability for formalized mathematics.
Specifically, our premise retrieval model consists of a context-free retrieval (CFR) module and a context-aware re-ranking (CAR) module, each of which takes BERT~\cite{bert_devlin2018bert} as its backbone.
The CFR module derives embeddings for premises and proof states (i.e., queries) and retrieves relevant premises based on their similarity to the input proof states.
The CAR module is employed to reorder the retrieved premises, and accordingly, improves the top-$k$ recall of the retrieval results. 
Unlike approaches that merely fine-tune pre-trained language models, we first pre-train a BERT model from scratch based on existing formalized corpora, in which a new tokenizer is learned for formal language.
Based on the pre-trained BERT, we further learn the CFR and CAR modules separately via contrastive learning~\cite{chen2020simple,he2020momentum}.
Experiments show that the retrieval results of our model surpass state-of-the-art models in performance while still maintaining computational efficiency. 
Furthermore, to evaluate the practicality of our premise retrieval model, we integrate it with an independently trained tactic generator, creating a retrieval-augmented automated theorem prover. 
We then assess this system using the test dataset and MiniF2F benchmark~\cite{zheng2021minif2f}, demonstrating the effectiveness of our retrieval model in facilitating the proof process.

Our contributions can be summarized as follows:
\begin{itemize}[leftmargin=*]
    \item We train an effective retrieval model for formalized theorems, with a new tokenizer for formal language, which achieves notable results at low computational costs and capable of running on a personal computer.
    \item We evaluate various retrieval models in a retrieval-augmented theorem proving pipeline, demonstrating the advantages of the proposed retrieval model in assisting proof generators.
    \item Based on our model, we have deployed a search engine featuring a real-time updating database, available free of charge to provers, and provide a toolkit to facilitate local model deployment for users.
\end{itemize}

%% file: sections/preliminaries.tex
\section{Preliminaries and Related Work}
\subsection{Preliminaries in Formalization and Lean}\label{preliminaries}

As a proof assistant, Lean provides a user interface for constructing formal proofs. 
The formalization process typically begins by translating a mathematical statement into Lean. 
Lean then analyzes the statement and presents the user with a proof state, which consists of a collection of hypotheses and the proposition to be proved. 
The user provides commands to modify the proof state until no goals remain.
In this context, the set of hypotheses is referred to as the \textit{context}, and the proposition is called the \textit{goal}.

While there are various styles of proofwriting, machine learning researchers commonly adopt the tactic-state transformation as an abstraction of the process. 
A tactic is a piece of program that modifies the current proof state. 
Experienced Lean users can develop custom tactics through meta-programming, utilizing Lean’s native automation tools. 
Pre-defined tactics, such as \verb|simp|, \verb|linarith|, and so on, have significantly enhanced the efficiency of formalization.\footnote{The \texttt{simp} tactic uses lemmas and hypotheses to simplify the main goal target or non-dependent hypotheses. The \texttt{linarith} tactic attempts to find a contradiction between hypotheses that are linear (in)equalities.}

\begin{figure}[t]
    \centering
    \includegraphics[width=0.9\linewidth]{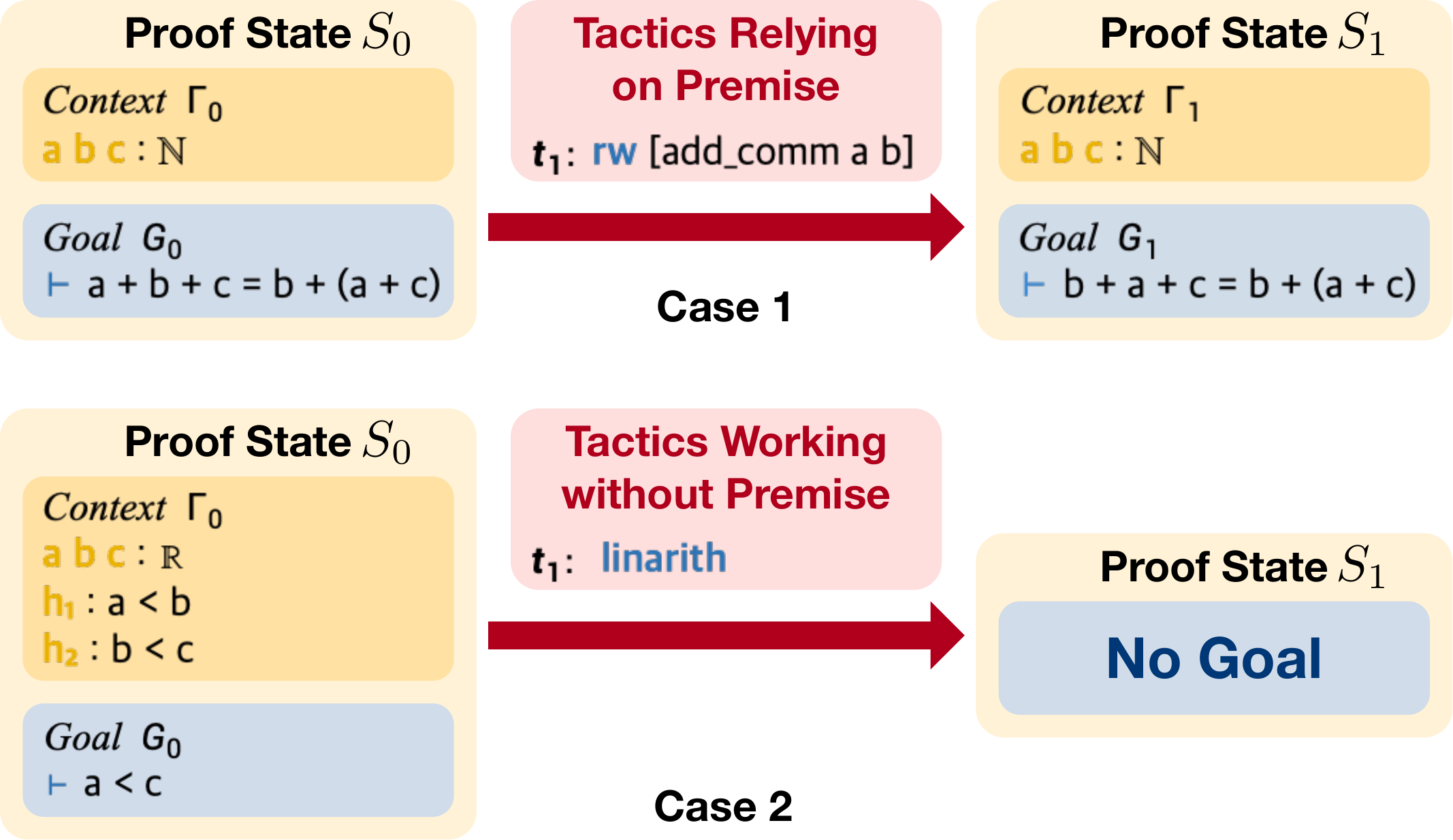}
    \caption{Examples of the theorem-proving process in Lean.}
    \label{fig:tactic}
\end{figure}

We present a formal specification of the theorem-proving process in Lean. 
Starting with an initial proof state $S_{0} = \{\Gamma_{0}, G_{0}\}$, where $\Gamma_{0}$ represents the initial context and $G_{0}$ the initial goal, the prover must provide a sequence of tactics $\{t_{i}\}_{i=1}^{n}$ to construct the proof. 
This sequence can be visualized as a linked chain of transformations, i.e., 
\begin{equation}\label{eq:seq}
    S_{0}\overset{t_{1}}{\longrightarrow}S_{1}\overset{t_{2}}{\longrightarrow}S_{2}\overset{t_{3}}{\longrightarrow}\cdots\overset{t_{n}}{\longrightarrow}S_{n}.
\end{equation}
The proof is complete when $G_{n} = \text{``No Goals''}$. 
Note that tactics are naturally various.
As shown in Figure~\ref{fig:tactic}, some require additional premises to take effect, while others operate independently. 
Furthermore, the behavior of certain tactics may vary depending on the presence of premises. 

%% file: sections/relatedwork.tex
\subsection{Related Work}

\textbf{Learning-based Retrieval Models} 
In the field of information retrieval, traditional methods like BM25~\cite{bm25_harman1995overview, bm25_robertson2009probabilistic} and TF-IDF~\cite{tfidf_salton1988term} rely on term information and document statistics to match queries with relevant documents. 
With the advancement of deep learning, researchers have developed learning-based retrieval models, often referred to as dense retrieval models~\cite{dense_passage_retrieval_karpukhin2020dense, dense_passage_retrieval_rocket_qu2020rocketqa}. 
These models typically fall into two categories.

The first approach~\cite{dense_passage_retrieval_karpukhin2020dense, dense_passage_retrieval_rocket_qu2020rocketqa} encodes queries and documents into embeddings separately.
Their similarity is then computed in the embedding space, and the top-ranked documents are retrieved as the final results. 
A key advantage of this method is that document embeddings can be precomputed, allowing the system to embed only the query during retrieval and improving computational efficiency significantly. 
However, this approach lacks direct interaction between queries and documents, which may limit its retrieval performance.
The second approach~\cite{cross_encoder_qiao2019understanding, rerank_nogueira2019passage} concatenates the query and document as a single input to the encoder model, which then outputs a relevance score between them.
This method can capture richer interactions between queries and documents, leading to improved performance. 
However, it incurs higher computational costs since the concatenation must be performed for each query-document pair.
Recently, researchers have increasingly embraced the paradigm of fine-tuning pre-trained models for various retrieval tasks~\cite{retrieval_task_ma2024fine, retrieval_task_wang2023improving, retrieval_task_li2024llama2vec}, leading to substantial improvements in performance.

\textbf{Premise Selection in Lean}
Premises are theorems or hypotheses that can be used in a tactic.
Lean's internal API\footnote{Loogle: \url{https://loogle.lean-lang.org/}} employs pattern matching to identify theorems applicable to the current proof state. 
This approach, however, enforces strict matching criteria, which often leads to failures in retrieving relevant theorems.
In addition, applications like Moogle\footnote{Moogle: \url{https://www.moogle.ai/}} and leansearch~\cite{pku-gao-etal-2024-semantic-search} have been developed to enable searches based on natural language descriptions of the target query.
However, they place a higher demand on the precision of natural language formulations for both the query and the formal theorems.
In contrast, premise selection based on purely current formal proof state is a more straightforward approach.
LeanDojo~\cite{lean_dojo_yang2024leandojo} employs a learning-based retrieval model that encodes the proof state and theorems independently, subsequently calculating their similarity to perform the retrieval task. 

%% file: sections/method.tex
\begin{figure*}[t]
    \centering
    \includegraphics[width=\linewidth]{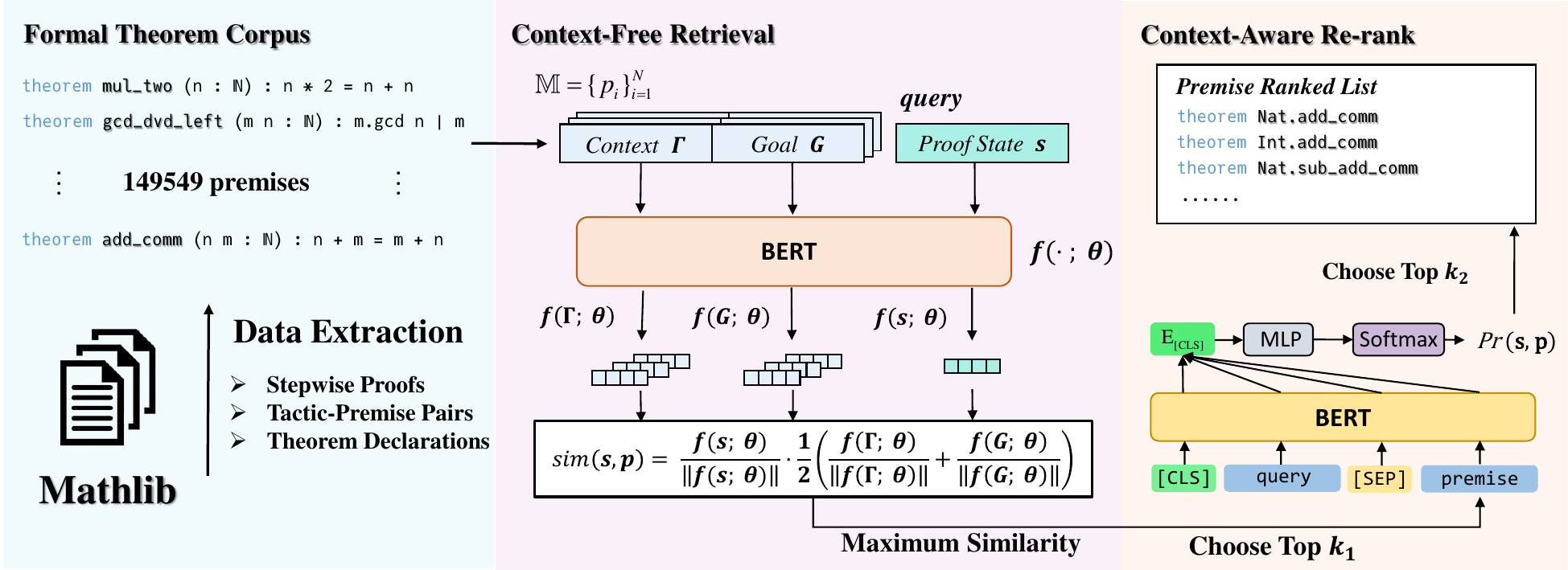}
    \caption{The overview of the retrieval framework. We extract 149,549 premises from Mathlib as our formal theorem corpus. In the context-free retrieval stage, all the premises are encoded by averaging the embeddings of context and goal. The top-$k_{1}$ premises retrieved in this stage will be re-ranked in a context-aware manner.}
    \label{fig:architecture}
\end{figure*}

\section{Proposed Method}

As shown in Figure~\ref{fig:architecture}, our work includes three stages:
1) Extract data from the formal mathematics library to construct the theorem corpus. 
2) The context-free retrieval stage involves encoding the premises and proof states using a learning-based proof state encoder and calculating semantic similarities. 
3) In the context-aware re-ranking stage, the retrieved premises are concatenated with the query and evaluated using a cross-encoder for classification and re-ranking.


\subsection{Data Preparation}\label{data preparation}

Let $\mathbb{M} = \{\boldsymbol{p}_i\}_{i=1}^N$ denote the library to be searched, where each $\boldsymbol{p}_i$ represents a theorem within the library. 
Since theorems are functions that accept hypotheses as input and yield proofs of specified propositions, $\boldsymbol{p}_i$ can be represented as $\{\Gamma_{\boldsymbol{p}_i}, G_{\boldsymbol{p}_i}\}$, where $\Gamma_{\boldsymbol{p}_i}=[v_1,\dots, v_n]_{\boldsymbol{p}_i}$ is the argument list of the theorem and $G_{\boldsymbol{p}_i}$ is the output of the theorem. 
This representation is isomorphic to a proof state. 
Given the current proof state $\boldsymbol{s}$, the parameterized retrieval model $\Phi(\boldsymbol{s}; \mathbb{M}, \Theta)$ assigns a score to each premise and returns the top-$k$ results. 
Note that a proof state may contain several cases, denoted as $\boldsymbol{s}=\{\boldsymbol{s}^{i}\}_{i=1}^{|\boldsymbol{s}|}$, where $\boldsymbol{s}^i=\{\Gamma_{\boldsymbol{s}}^i, G_{\boldsymbol{s}}^i\}$.

From human-written proofs, we extract tactic steps $\mathbb{T} = \{t_i\}_{i=1}^T$. 
Each tactic step contains the proof state before and after executing the tactic, as well as the premises used in the tactic, denoted as $t_i = (\boldsymbol{s}_i^{\rm{before}}, \{\boldsymbol{p}_{ij}\}_{j=1}^{n_i}, \boldsymbol{s}_i^{\rm{after}})$, where $\boldsymbol{p}_{ij} \in \mathbb{M}$. We collect state-premise pairs from these tactics to form our training dataset. 
We consider the state before/after a tactic both relevant to the premises used in this step since the state before should satisfy the hypotheses of the premises, while the state after will contain patterns in the goal of the premises. 
Thus, we form our dataset as $\mathcal{D}=\{(\boldsymbol{s}_i, \mathcal{P}_i)\}_{i=1}^{|\mathcal{D}|}$, where $\boldsymbol{s}_i$ are distinct proof states and $\mathcal{P}_i$ is the set of relevant premises of $\boldsymbol{s}_i$. 
\subsection{Model Architecture}\label{model architecture}
Our premise retrieval model consists of two components: a context-free retrieval (CFR) module and a context-aware re-ranking (CAR) module.
In detail, the CFR module retrieves a large set of relevant theorems for a given proof state. 
Subsequently, the CAR module refines the ranking of the top-$k$ theorems retrieved by the CFR module, further improving the accuracy of the retrieval process.

\textbf{Context-Free Retrieval}
In the context-free retrieval stage, we adopt a dense retrieval method \cite{dense_passage_retrieval_karpukhin2020dense}. 
This approach maps both the proof state and the theorem into a shared latent space, enabling retrieval through the computation of their cosine similarity. 
We design the model based on the BERT encoder architecture. 
For a given input text, we process it through the model and apply average pooling over the last hidden states to derive its latent representation, denoted as \( f(\cdot; \theta) \).

We preprocess the context and goal for each state and premise to eliminate the effect of the theorem name and achieve a unified format. 
We prepend a special token \verb|<VAR>| to each element of the context $\Gamma$ and \verb|<GOAL>| to the goal $G$, then concatenate them to form a new string. Here is an example:

\begin{tcolorbox}[colframe=black, colback=white]
Raw theorem:
\begin{leancode}
theorem Nat.add_comm (n m : Nat) : 
    n + m = m + n 
\end{leancode}
After process:
\begin{leancode}
<VAR> n m : Nat <GOAL> n + m = m + n
\end{leancode}
\end{tcolorbox}

The retrieval model measures the relevance between a proof state $\boldsymbol{s}$ and a premise $\boldsymbol{p}$ by computing the similarity of their embeddings. 
The conventional similarity can be defined as
\begin{equation}\label{eq:gloabl_sim}
    \text{sim}(\boldsymbol{s}, \boldsymbol{p})=\frac{f(\boldsymbol{s};\theta)}{\|f(\boldsymbol{s};\theta)\|}\cdot\frac{f(\boldsymbol{p};\theta)}{\|f(\boldsymbol{p};\theta)\|}.
\end{equation}
However, considering that some states may only correlate with either the arguments or the goal, we encode the premise arguments and goal separately, which can be denoted as $f(\Gamma_{\boldsymbol{p}}, \theta)$ and $f(G_{\boldsymbol{p}}, \theta)$.
Accordingly, we design a fine-grained similarity between the given state $\boldsymbol{s}$ and premise $\boldsymbol{p}$ --- $\text{sim}(\boldsymbol{s}; \boldsymbol{p})$ is formulated as
\begin{equation}
     \underbrace{\frac{f(\boldsymbol{s};\theta)}{\|f(\boldsymbol{s};\theta)\|}}_{\text{state embedding}}\cdot \underbrace{\frac{1}{2} \left(\frac{f(\Gamma_{\boldsymbol{p}}; \theta)}{ \|f(\Gamma_{\boldsymbol{p}}; \theta)\|} + \frac{ f(G_{\boldsymbol{p}}; \theta)}{ \|f(G_{\boldsymbol{p}}; \theta)\|}\right)}_{\text{premise embedding}}.
\label{eq:similarity}
\end{equation}
This similarity is designed based on the principle that the context of the proof state should first meet the preconditions of the available premises and, afterward, determine whether the conclusions drawn from these premises fulfill the current requirements. 
Compared to the conventional similarity, this fine-grained design aims to explicitly measure the similarity from both perspectives, capturing premises that can be applied in the current context, along with premises that can be a solution to the current goal.

 

\textbf{Context-Aware Re-ranking}
While facilitating the retrieval model and vector store enables fast and efficient queries, the lack of interaction between the state and premise may lead to lower accuracy. 
To address the limitations of this context-free approach, we adopt a context-aware method to re-ranking the results from first stage. 
Specifically, we design the re-ranking model based on BERT architecture that follows the sequence-pair classification of~\cite{rerank_nogueira2019passage}.
The state and premise will be concatenated together by \verb|[SEP]| as the re-ranking module's input sequence.

The module obtains the embeddings of the input. 
Passing the \verb|[CLS]| embedding through an affine projection followed by a sigmoid layer, we obtain the relevant probability of the state and premise, i.e., 
\begin{equation}\label{eq:seq}
Pr(\boldsymbol{s}, \boldsymbol{p}) = \sigma(\mathbf{W} \cdot \mathbf{h}_{[CLS]} + \mathbf{b}),
\end{equation}
where $\mathbf{h}_{[CLS]}$ is the embedding of the \verb|[CLS]| token, $\mathbf{W}$ is the weight matrix, $\mathbf{b}$ is the bias term, and $\sigma$ represents the sigmoid activation function.
For top-$k$ retrieval, we first use the retrieval model to find $k_1$ candidates. 
The re-ranking model serves as a filter and returns top-$k$ as the final result.

Note that, existing re-ranking methods~\cite{cross_encoder_qiao2019understanding, rerank_nogueira2019passage} improve retrieval performance by concatenating the query and each passage and passing them through their re-ranking models. 
This strategy has a high computational cost during inference due to the combining process of state and premise. 
As a trade-off, we apply a re-ranking model to refine the ranking of premises retrieved by the retrieval model.

\subsection{Learning Algorithm}\label{learning algorithm}

Based on the Masked Language Modeling (MLM)~\cite{MLM_taylor1953cloze, bert_devlin2018bert} method, we pre-train the CFR and CAR modules on the formalized corpus we collected.
Their tokenizers are the same and trained on formalized language corpus using WordPiece~\cite{wordpiece2021xin} algorithm. 
Then, we use the state-premises pairs to fine-tune the pre-trained modules separately by contrastive learning.

\textbf{Contrastive Learning of CFR Module}
\label{sec:contrastive learning}
For $(\boldsymbol{s}_i, \mathcal{P}_i)\in \mathcal{D}$, we use $\boldsymbol{p}_{ij} \in \mathcal{P}_i$ along with negatives sampled from $\mathbb{M}$ to construct a set $\mathcal{P}_{ij}'$, which contains one positive $\boldsymbol{p}_{ij}$ and $|\mathcal{P}_{ij}'|-1$ negatives.
The optimization objective of contrastive learning is formulated as
\begin{equation}
\min \sum_{i=1}^{|\mathcal{D}|}\sum_{j=1}^{|\mathcal{P}_i|} -\log \frac{\exp(\text{sim}(\boldsymbol{s}_{i}, \boldsymbol{p}_{ij})  / \tau)}{\sum_{\boldsymbol{p}' \in \mathcal{P}_{ij}'} \exp(\text{sim}(\boldsymbol{s}_i,\boldsymbol{p}') / \tau)},
\label{eq:contrastive_loss}
\end{equation}
where $\tau$ refers to the temperature. 
We use the Homogeneous In-Batch Negative Sampling, which calls for many negative samples to guarantee the embedding's discriminativenss~\cite{discriminative_izacard2021unsupervised, discriminative_wang2022text, dense_passage_retrieval_rocket_qu2020rocketqa}.
In our work, this is implemented by the usage of in-batch negatives --- for each query, the negatives are randomly sampled from the corpus, excluding the other positive premises of the query. 
Given a batch of $B$ samples, it results in $B\times |\mathcal{P}_{ij}'|-1$ negative samples.

\textbf{Contrastive Learning of CAR module} 
The CAR module is learned in the same contrastive learning framework, in which the choice of negatives is crucial. 
We sample hard negatives from the top-$k_{1}$ premises selected by the retrieval model and they will be used in the re-ranking model training process.
During testing, we also use the retrieval model to retrieve the top-$k_{1}$ premises and re-ranking them by the re-ranking model. 
The loss function used for training the re-ranking model is the cross-entropy loss, and the optimization objective is formulated as
\begin{equation}
\min \sum_{i=1}^{|\mathcal{D}|} \sum_{j=1}^{|\mathcal{P}_i|} -\log \frac{Pr(\boldsymbol{s}_i, \boldsymbol{p}_{ij})}{\sum_{\boldsymbol{p}' \in \mathcal{P}_{ij}'} Pr(\boldsymbol{s}_i, \boldsymbol{p}')},
\label{eq:contrastive_loss2}
\end{equation}
where $\mathcal{P}_{ij}'$ contains one positive $\boldsymbol{p}_{ij}$ and some hard negatives.
$Pr(\boldsymbol{s}, \boldsymbol{p})$ is the relevant probability obtained via Eq.~\eqref{eq:seq}.

\textbf{Training Tactic Generator} 
\label{sec:tactic generation}
After training the retrieval model, we integrate it with the generator for testing. 
The generator is trained in an independent paradigm for a fair comparison between retrievers. 
Most prior work on generators~\cite{llmstep_welleck2023llmstep, lean_copilot_song2024towards, lean_dojo_yang2024leandojo} utilizes state-tactic pairs to fine-tune a pre-trained model for tactic generation.
Leandojo~\cite{lean_dojo_yang2024leandojo} incorporates premise retrieval by leveraging the retrieval model to extract relevant premises based on the state from the training dataset.
The premises are then concatenated with the state to form premise-augmented state-tactic pairs used for fine-tuning.
However, this approach results in a high degree of coupling between the retrieval model and the generator, as prior knowledge of the retrieval result distribution is incorporated during training.

To ensure a fair experimental setup, we aim to decouple the generator from the retrieval model and train the generator independently.
As a substitution of the retrieval model, we randomly select the positive and negative premises from the library and prepend them to the state, thereby creating premise-augmented state-tactic pairs.
For each tactic, we add up to ten premises. 
The number of positive premises selected is randomly determined, ranging from one to the lesser of the total available positive premises or ten. 
The remaining premises are negative. 
Half of these negatives are selected from the same module as the positive samples, while the remaining are chosen from other modules.
Following~\cite{lean_dojo_yang2024leandojo}, we use these pairs to fine-tune the ByT5~\cite{Byt5_xue2022byt5} model for tactic generation.
The loss function we use is typically cross-entropy loss.

%% file: sections/experiments.tex
\section{Experiments}

\subsection{Experiment Setup}\label{implement details}

\textbf{Implementation Details}
Our retrieval and re-ranking models are based on the BERT architecture, which consists of 6 layers, each with 12 attention heads. 
It features a hidden size of 768 and an intermediate size of 3,072, complemented by a vocabulary size of 30,522.
For the retrieval model, the maximum position embeddings are configured at 512.
The maximum length for states is set at 512 and for premises' arguments and goals set at 256 respectively. 
The batch size is set at 32, and $|\mathcal{P}_{ij}'|$ mentioned in~\ref{sec:contrastive learning} is set at 2. 
In contrast, the re-ranking model has its maximum position embeddings set at 1,024. 
We set the batch size at 2, gradient accumulation steps at 8, and $|\mathcal{P}_{ij}'|$ at 8.
We construct the pre-training corpus by concatenating states from the training set and all premises' argument lists and goals from the corpus. 
The results of our method presented in Table~\ref{tab:main result} are obtained by first training the retrieval model and selecting the top-$100$ results as hard negative candidates. 
We then train the re-ranking model and use it to reorder the top-$20$ results from the retrieval stage.
The experiments are conducted on 8 RTX 4080 servers.

\textbf{Baselines}
\label{sec:baseline}
As mentioned above, ReProver~\cite{lean_dojo_yang2024leandojo} is a model specifically trained on the Lean dataset, using formal states and premises. 
Therefore, it serves as our primary baseline. 
Moreover, we retrain the model using our dataset and following the setting in~\cite{lean_dojo_yang2024leandojo}.
In addition to ReProver, we compare against several other commonly used retrieval models 
that have demonstrated strong performance in their specific domains.
\begin{itemize}[leftmargin=*]
    \item UniXcoder-base~\cite{unixcoder_guo2022unixcoder}: It is one of the state-of-the-art code embedding models to transform code snippets into hidden vectors. 
    It leverages multimodal data, such as code comments and abstract syntax trees (ASTs), to pre-train code representations.
    \item E5-large-v2~\cite{discriminative_wang2022text}: It is trained in a contrastive manner using weak supervision signals derived from a curated, large-scale text-pair dataset (referred to as CCPairs) and demonstrates strong performance on several English retrieval tasks.
    \item BGE-m3~\cite{bge-m3-multim3}: It is distinguished for its versatility and excels in multiple functionalities, multilingual capabilities, and fine granularity retrieval tasks.
\end{itemize}

We fine-tune the three aforementioned baselines using our dataset. 
Following the parameters provided in~\cite{bge-m3-multim3}, we set the learning rate to 1e-5 and use a linear scheduler. 
Due to server constraints, the batch sizes for the three models are set to 16, 8, and 6, respectively.

\textbf{Metrics}
To evaluate the performance of various retrieval models, we utilize four widely adopted metrics: Precision, Recall, F1-Score, and nDCG. 
Since the nDCG considers the retrieved results' relevance and position, it allows for multiple relevance levels.
Following the setup in~\cite{pku-gao-etal-2024-semantic-search}, we define the relevance criteria in Table~\ref{tab:relevance criteria}, which provides the relevance scores for calculating nDCG. 

\begin{table}[ht]
    \centering
    \caption{Relevance criteria for the retrieval results.}
    \label{tab:relevance criteria}
    \begin{small} 
    \def\arraystretch{1.5}
    \tabcolsep=1pt
    \begin{tabular}{c|c|m{6cm}}
    \hline\hline \xrowht{5pt}
    \textbf{Rating} & \textbf{Score} & \textbf{Description} \\
    \hline
    Match & 1 & Match to one of the premises used in the tactic. \\
    \hline
    Relevant & 0.3 & Within the same module with any of the premises used in the tactic. \\
    \hline
    Irrelevant & 0 & Situations other than the two mentioned above. \\
    \hline\hline
    \end{tabular}
    \end{small}
\end{table}

\textbf{Data Split}
To evaluate the performance of the model across different feature datasets, we employ four distinct data split strategies.
Here, we represent a proof as $P = \{t_i\}_{i=1}^{|P|}$, which is a sequence of tactics, and apply \textit{Proof Length} and \textit{Premise Frequency} to characterize the proof, i.e., 
\begin{eqnarray*}
\begin{aligned}
\mathrm{ProofLength}&=|P|,   \\ \mathrm{PremiseFreq}&=\frac{1}{|P|}\sideset{}{_{i=1}^{|P|}}\sum\mathrm{PremiseNum}(t_i),
\end{aligned}
\end{eqnarray*}
where $\mathrm{PremiseNum}(t_i)$ denotes the number of external premise in a tactic. 
\textit{Proof Length} reflects the complexity of the proof, while the \textit{Premise Frequency} indicates the degree of reliance on external theorems during the proof process. 
Then, we apply the following four data split strategies to obtain four datasets:
\begin{itemize}[leftmargin=*]
    \item \textbf{Random (RD):} Randomly split the dataset.
    \item \textbf{Reference Isolated (RI):} Premises in the validation and test sets will not appear in the training set.
    \item \textbf{Proof Length (PL):} Sample proofs for the validation and test sets weighted by \textit{Proof Length}.
    \item \textbf{Premise Frequency (PF):} Sample proofs for the validation and test sets weighted by \textit{Premise Frequency}.
\end{itemize}
Each dataset contains 65,567 theorems for training, while the validation and test sets each contain 2,000 theorems. 


\begin{table*}[ht]
    \centering
    \caption{Premise selection results of our method and other baselines on different data splits.}
    \label{tab:main result}
    \resizebox{\linewidth}{!}
    {\begin{tabular}{c|l|ccc|ccc|ccc|ccc}
    \hline\hline
    \multirow{2}{*}{Split} & \multicolumn{1}{c|}{\multirow{2}{*}{Method}} & \multicolumn{3}{c|}{Recall}&\multicolumn{3}{c|}{Precision}&\multicolumn{3}{c|}{F1-score} & \multicolumn{3}{c}{nDCG}\\ 
    & & R@1 & R@5 & R@10 & P@1 & P@5 & P@10 & F@1 & F@5 & F@10 & n@1 & n@5 & n@10 \\ \hline
    \multirow{5}{*}{RD} &UniXcoder-base & 4.23 & 19.47 & 27.01 & 6.68 & 6.90 & 5.04 & 5.18 & 10.19 & 8.50 &  0.2283 
 &  0.3455 
&  0.3958 
 \\
    & E5-large-v2 &4.42 &18.99 &27.29 &6.90 &6.75 &5.09 &5.39 &9.96 &8.58 & 0.2313 
 & 0.3458 
 & 0.3952 
 \\
    & BGE-m3 &4.00 &18.56 &25.29 &6.40 &6.51 &4.64 &4.92 &9.64 &7.84& 0.2374 
 & 0.3481 
 & 0.3930 
 \\
    & ReProver & 11.79 & 28.78 & 36.69 & 19.80 & 10.90 & 7.23 & 14.78 & 15.81 & 12.08 &   0.3351 
 
 &   0.4072 
 
 &   0.4617 
 
 \\
    & Ours & \textbf{15.17} & \textbf{38.20} & \textbf{46.53} & \textbf{26.80} & \textbf{14.87} & \textbf{9.30} & \textbf{19.38} & \textbf{21.41} & \textbf{15.51} & \textbf{0.3731} & \textbf{0.4698} & \textbf{0.5163} \\ 
    \hline
    \multirow{5}{*}{RI} &UniXcoder-base & 4.54 & 17.55 & 24.16 & 6.68 & 6.18 & 4.54 & 5.41 & 9.15 & 7.64 &  0.2303 
 &  0.3258 
 &  0.3768 
 \\
    &E5-large-v2 & 4.80 & 18.21 & 24.73 & 7.42 & 6.55 & 4.69 & 5.83 & 9.64 & 7.88 &  0.2356 
& 0.3407 
 & 0.3878 
 \\
    &BGE-m3 & 4.52 & 17.69 & 23.76 & 7.05 & 6.19 & 4.45 & 5.51 & 9.17 & 7.49 & 0.2306 
 & 0.3312 
 & 0.3788 
 \\
    &ReProver & 5.05 & 14.26 & 19.48 & 9.30 & 5.77 & 4.10 & 6.54 & 8.22 & 6.77 & 0.2442 
 & 0.3059 
 & 0.3563 
 \\
    & Ours & \textbf{7.79} & \textbf{23.38} & \textbf{30.91} & \textbf{14.76} & \textbf{9.30} & \textbf{6.38} & \textbf{10.20} & \textbf{13.31} & \textbf{10.58} & \textbf{0.2731} & \textbf{0.3736} & \textbf{0.4322} \\ 
    \hline
    \multirow{5}{*}{PL} & UniXcoder-base & 3.94 & 15.45 & 22.72 & 6.70 & 5.66 & 4.32 & 4.96 & 8.29 & 7.25 & 0.1958 
& 0.2804 
& 0.3272 
\\
    & E5-large-v2 & 3.66 & 14.65 & 21.65 & 6.25 & 5.47 & 4.23 & 4.61 & 7.96 & 7.08 & 0.1849 
& 0.2735 
& 0.3229 
\\
    & BGE-m3 & 3.99 & 14.65 & 20.53 & 7.06 & 5.41 & 3.92 & 5.10 & 7.90 & 6.59 & 0.1974 
& 0.2781 
& 0.3218 
\\
    & ReProver & 11.16 & 26.83 & 33.96 & 19.64 & 10.31 & 6.83 & 14.23 & 14.90 & 11.38 & 0.3059 
& 0.3699 
& 0.4232 
\\
    & Ours & \textbf{14.39} & \textbf{34.99} & \textbf{41.61} & \textbf{25.18} & \textbf{13.80} & \textbf{8.45} & \textbf{18.31} & \textbf{19.80} & \textbf{14.04} & \textbf{0.3479} & \textbf{0.4319} & \textbf{0.4781} \\ 
    \hline
    \multirow{5}{*}{PF} &UniXcoder-base & 2.69 & 15.03 & 21.91 & 5.19 & 6.92 & 5.43 & 3.54 & 9.48 & 8.70 & 0.2251 
 & 0.3426 
 & 0.3973 
\\
    & E5-large-v2 & 3.08 & 14.44 & 22.11 & 5.84 & 6.70 & 5.39 & 4.04 & 9.16 & 8.67 & 0.2300 
& 0.3464 
& 0.4023 
\\
    & BGE-m3 & 2.59 & 13.85 & 20.05 & 5.42 & 6.27 & 4.86 & 3.51 & 8.63 & 7.83 & 0.2320 
& 0.3404 
& 0.3910 
\\
    & ReProver & 8.29 & 22.74 & 30.21 & 18.99 & 11.83 & 8.30 & 11.54 & 15.57 & 13.02 & 0.3257 
& 0.3966 
& 0.4544 
 \\
    & Ours & \textbf{11.44} & \textbf{31.02} & \textbf{38.88} & \textbf{26.38} & \textbf{15.96} & \textbf{10.52} & \textbf{15.96} & \textbf{21.08} & \textbf{16.56} & \textbf{0.3764} & \textbf{0.4693} & \textbf{0.5238} \\ 
    \hline \hline
    \end{tabular}}
\end{table*}

\begin{table*}[t]
    \centering
     \caption{Ablation studies on pre-training, tokenizer, and similarity calculation on Random split dataset for the CFR module. Here P means the model is pre-trained or not, T means the model uses a new tokenizer or not, S means similarity calculation.}
     
    \label{tab:pretrain_token}
    \begin{small}
    \begin{tabular}{ccc|ccc|ccc|ccc|ccc}
    \hline\hline
    \multicolumn{3}{c|}{Method} & \multicolumn{3}{c|}{Recall} & \multicolumn{3}{c|}{Precision} & \multicolumn{3}{c|}{F1-score} &  \multicolumn{3}{c}{nDCG} \\ 
    P& T& S & R@1 & R@5 & R@10 & P@1 & P@5 & P@10 & F@1 & F@5 & F@10 & n@1& n@5 & n@10 \\ \hline
    $\times$ &$\times$&Eq.~\eqref{eq:similarity}  & 4.63 & 20.09 & 29.05 & 8.35 & 7.45 & 5.62 & 5.96 &  10.87 
 &  9.42 
 & 0.2443 
 & 0.3503 
 & 0.4030 
 \\
    $\times$ & $\checkmark$& Eq.~\eqref{eq:similarity} & 4.93 & 20.49 & 30.33 & 8.04 & 7.50 & 5.79 & 6.11 &  10.98 
 &  9.72 
 & 0.2368 
& 0.3493 
& 0.4034 
 \\
    $\checkmark$ & $\times$& Eq.~\eqref{eq:similarity} & \textbf{6.31} & 23.81 & 34.21 & \textbf{10.34} & 8.77 & 6.59 & {7.84} &  12.82 
 &  11.05 
 & \textbf{0.2484}
 &  0.3655 
 & 0.4240 
\\
    $\checkmark$ & $\checkmark$& Eq.~\eqref{eq:gloabl_sim} & 5.65 & \textbf{25.87} & 36.78 & 9.36 & \textbf{9.55} & 7.12 & 7.05 &  \textbf{13.95} 
 &  11.92 
 & 0.2382 
&  0.3662 
& 0.4238 
\\
    $\checkmark$ & $\checkmark$& Eq.~\eqref{eq:similarity} & 5.94 & {25.81} & \textbf{37.70} & 9.80 & {9.53} & \textbf{7.24} & 7.40 &  13.92 
 &  \textbf{12.15} 
 & 0.2422 
& \textbf{0.3674} 
& \textbf{0.4293} 
\\

    \hline\hline
    \end{tabular}
    \end{small}
    
\end{table*}

\subsection{Main Results}\label{main results}
After training different models on the training dataset and evaluating them on the corresponding test sets, we obtain the results presented in Table~\ref{tab:main result}. 
From this table, we can observe that our method demonstrates exceptional performance, outperforming other baseline models. 
Specifically, UniXcoder-base~\cite{unixcoder_guo2022unixcoder}, E5-large-v2~\cite{discriminative_izacard2021unsupervised}, and BGE-m3~\cite{bge-m3-multim3} exhibit generally poor performance, which can be attributed to the substantial discrepancy between the pre-training corpora used by these models (based on natural language data) and the Lean language. 
Although these models are fine-tuned, they cannot fully capture the core aspects of the Lean language due to Lean's unique syntax and semantic features and thus suffer suboptimal performance. 
In contrast, the ReProver~\cite{lean_dojo_yang2024leandojo} model performs relatively well, likely due to its specifically designed parameter-tuning strategy.
Furthermore, ReProver utilizes ByT5~\cite{Byt5_xue2022byt5}, a byte-level tokenizer, which may allow the model to better handle the special symbols and structures in the Lean language. 
Our method, on the other hand, improves performance by retraining the tokenizer on the Lean corpus and pre-training the retrieval model, allowing it to better adapt to the syntax and semantic features of Lean.

For different data splitting strategies, we observe that random data splitting is relatively easy. 
However, when the theorems in the test set involve longer proofs or require a larger number of premises, the performance of the model tends to be lower. 
The reference isolated method, where the premises in the test set have not appeared during training, is the most challenging. 
It requires the model to exhibit strong generalization ability. 
Nevertheless, under each of these splitting strategies, our model consistently outperforms other baseline methods, which indicates that our model exhibits superior adaptability.

\subsection{Ablation Studies} \label{ablation study}


\begin{figure}[t]
    \centering
    \includegraphics[width=0.95\linewidth]{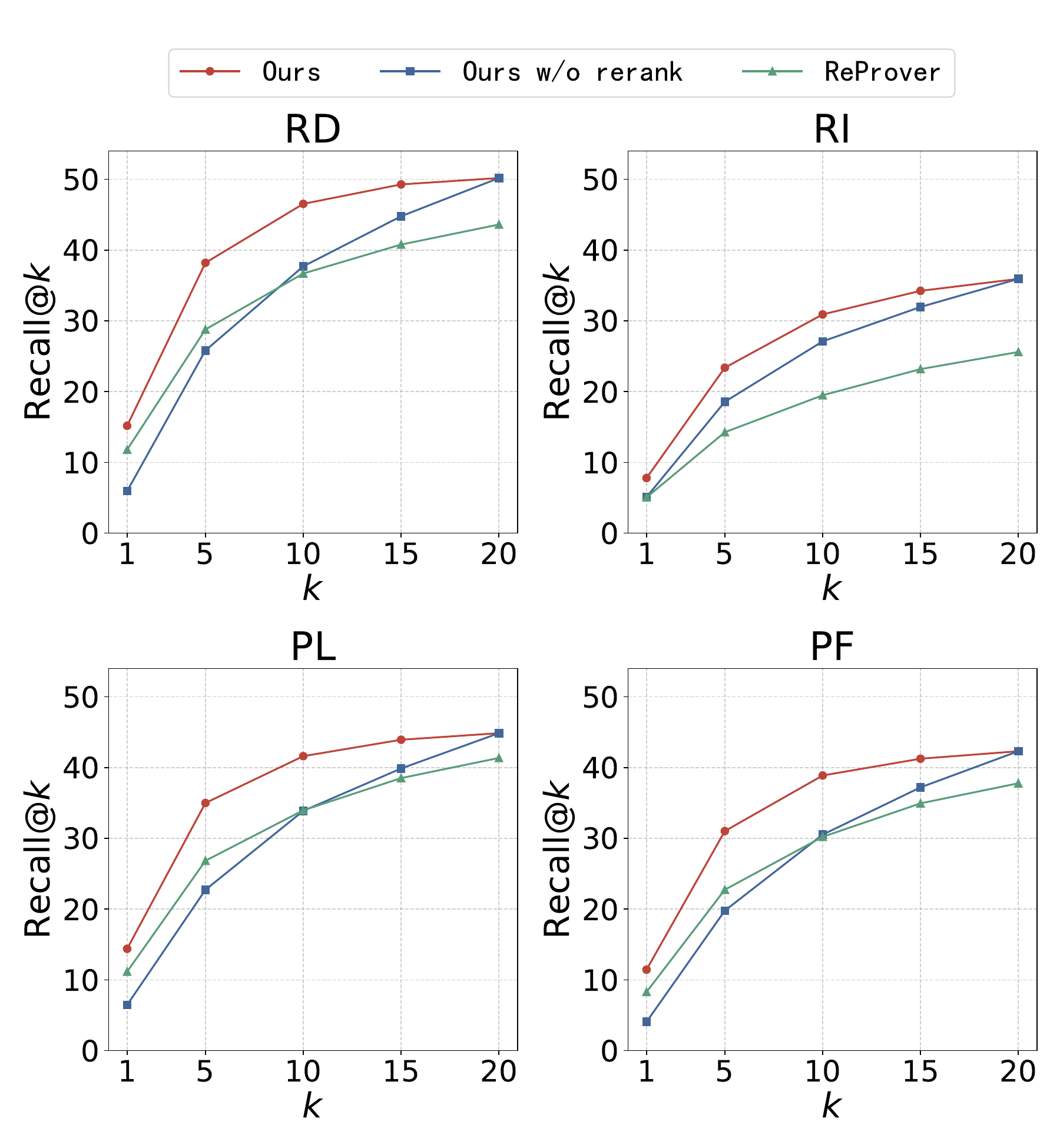}
    \caption{The Recall@k of our method with or without re-ranking on each split, compared with ReProver.}
    \label{fig:rerank_performance_a}
\end{figure}
\textbf{Impacts of Pre-Training, Tokenizer, and Similarity Calculation}
We evaluate the impacts of pre-training the model, training the specific tokenizer for Lean, and the proposed similarity calculation in Table~\ref{tab:pretrain_token}, respectively.
To ensure a fair comparison, we filter out the impact of the re-ranking module and only compare the results provided by the retrieval module. 
This table shows a performance improvement when the model is pre-trained on the Lean corpus compared to not performing pre-training. 
This indicates that the pre-training process helps the model acquire semantic information about Lean in advance. 
The experiment also shows that if we only pre-train the model without training a new tokenizer, the performance when $k=1$ is slightly better but degrades a lot when $k=5$ or $10$.
In addition, we evaluate the impacts of different similarity calculation methods on the retrieval results. 
The results in Table~\ref{tab:pretrain_token} emphasize the benefits of assessing similarity in a fine-grained manner.

\begin{figure}[h]
    \centering
    \includegraphics[width=0.8\linewidth]{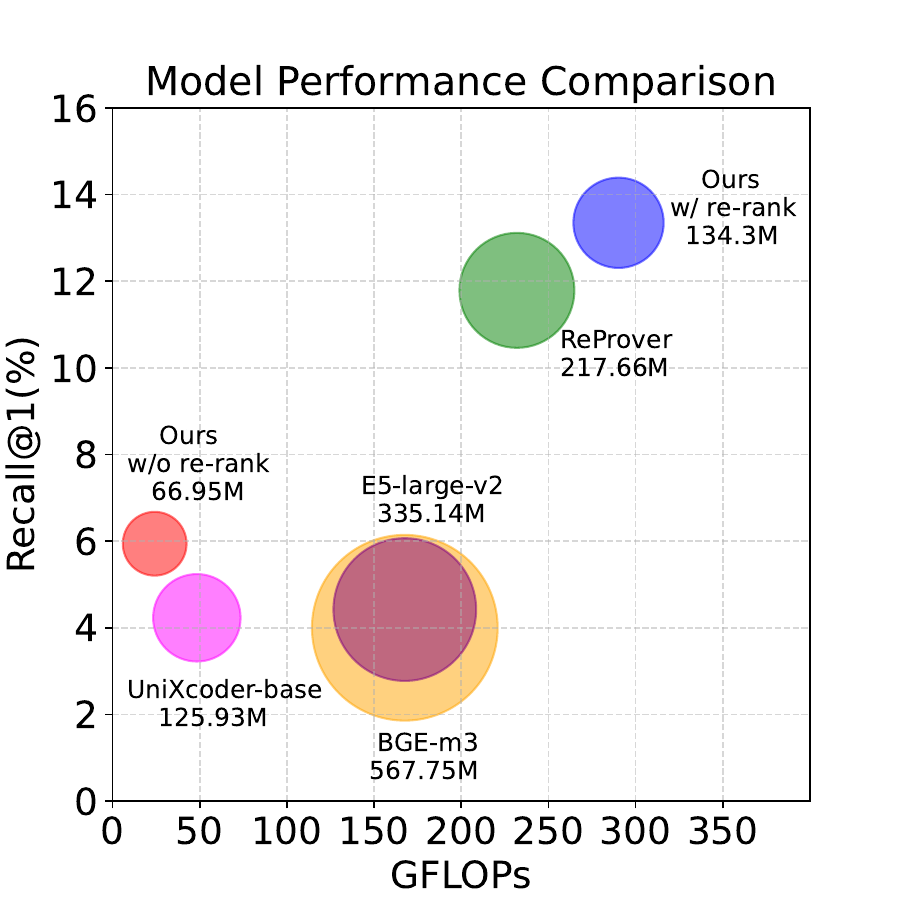}
    \caption{The performance and efficiency comparison on Random split datasets. The size of each point reflects the parameter size of each model.}
    \label{fig:rerank_performance_b}
\end{figure}

\textbf{The Effect of Re-ranking}
We validate the effectiveness of using the proposed re-ranking module. 
As shown in Figure~\ref{fig:rerank_performance_a}, when only the CFR module is used, our model outperforms the ReProver only for larger values of $k$ in Recall@k (e.g., $k\geq 10$). 
Applying the CAR module improves the recall of our model when $k<10$, so that our model can consistently outperform ReProver and users are likely to find useful premises in a relatively short list. 
Figure~\ref{fig:rerank_performance_b} presents a comparison for our model and other baselines in terms of model size and inference GFLOPs.
Without re-ranking, our model achieves the lowest GFLOPs and it outperforms in Recall@1 most of the baselines except ReProver.
When re-ranking the top-$5$ retrieval results by our CAR module, our model outperforms all the baselines and its GFLOPs and model size are comparable to those of ReProver.

\subsection{Robustness Experiments}

\begin{figure}[h]
    \centering
    \includegraphics[width=0.9\linewidth]{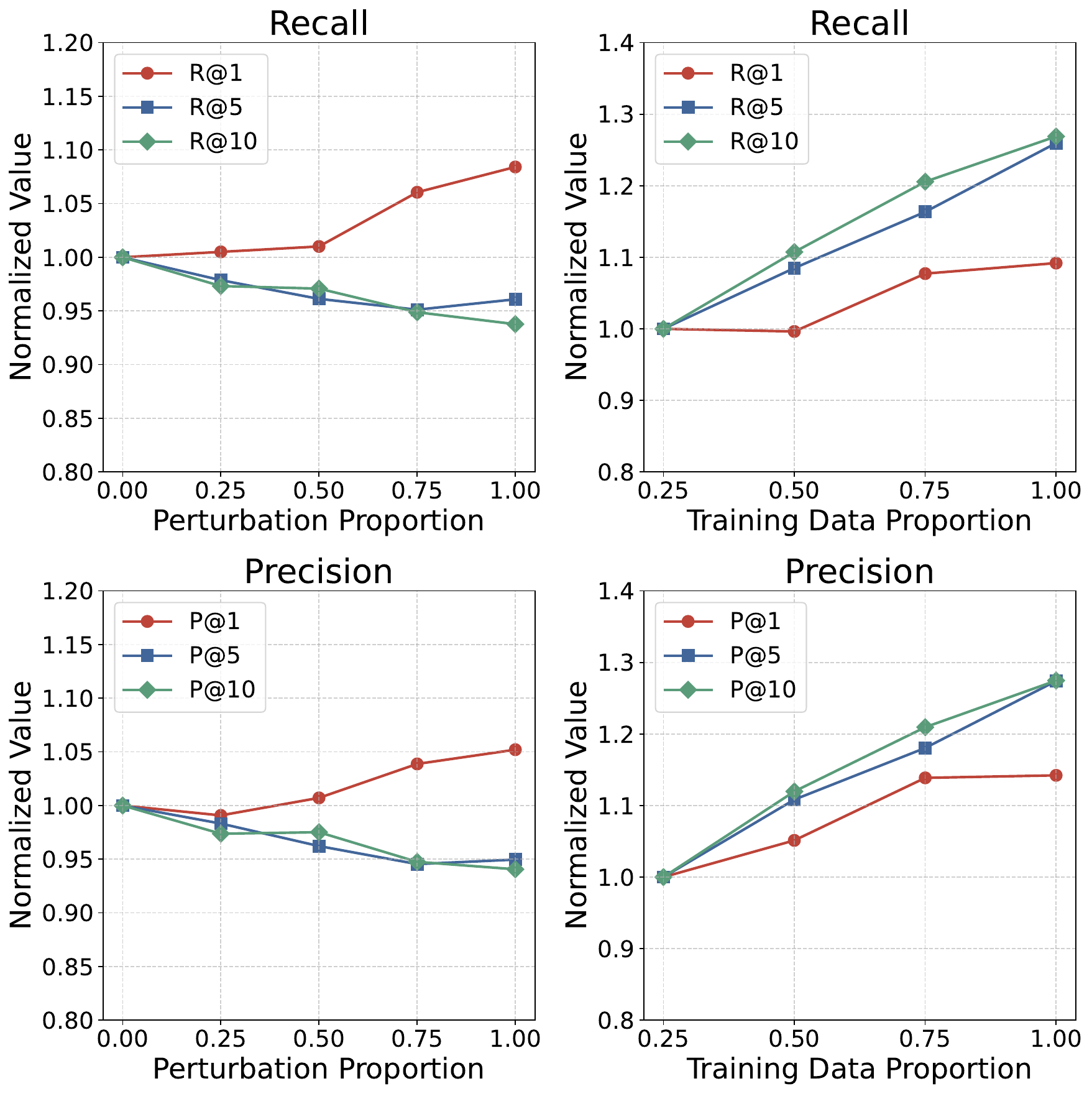}
    \caption{The variation of the normalized values of the model's metrics under different data perturbations and different training data proportion. 
    In data perturbation experiments, the metric with no perturbation $v_0$ is adopted as baseline. 
    The normalized value $v^*=v/v_0$. 
    In data proportion experiments, the metric with 25\% training data is adopted as baseline.}
    \label{fig:robustness}
\end{figure}

\textbf{Effect of Data Perturbation}
Considering the generally high quality of proofs in Mathlib, models trained on this dataset may be sensitive to the common issues of redundant variables, missing conditions, and disordered sequences in the context of proof states, resulting in poor generalization ability. 
We evaluate our retrieval model's robustness to low-quality inputs by perturbing the query states in the test set. 
Specifically, we applied two perturbation strategies to a subset of the test data: shuffling the context or randomly removing 20\% of the context from states with a context length of 15 or more. 
As shown in Figure~\ref{fig:robustness}, all the metrics exhibit a moderate decline as the perturbation ratio increases, with a maximum drop of approximately 6\% when $k=5,10$ compared to the unperturbed case.
Especially, the model's performance even improves when $k=1$. 
These results demonstrate that our retrieval model is robust to variations in the order of query states and the potential absence of certain local hypotheses.

\textbf{Impact of Reduced Training Data}
The sparsity of data is an inherent challenge in formalization tasks. 
We test the sensitivity of our retrieval model to the amount of training data. 
The tests are conducted on the Random split dataset.
The results in Figures~\ref{fig:robustness} show that the model trained on the full dataset outperforms the model trained on only 25\% of the data by approximately 25\% in terms of $k=5,10$.
The consistent performance improvement suggests that there is still potential for further enhancement by increasing the amount of training data.
When $k=1$, the performance seems to reach a bottleneck as the data size increases.
Therefore, it may be necessary to incorporate re-ranking or explore alternative methods to further improve performance.

\subsection{Theorem Proving Experiments}
As mentioned in Section~\ref{sec:tactic generation}, we fine-tune ByT5~\cite{Byt5_xue2022byt5} and obtain a tactic generator.
Facilitating the generator, we conduct theorem-proving experiments with retrieval. 
The results in Figure \ref{fig:prover result} show that assisted by our model, the generator averagely performs better than ReProver~\cite{lean_dojo_yang2024leandojo}.
Additionally, we evaluate our method on the Minif2f benchmark~\cite{zheng2021minif2f} using the model trained on the random split dataset.
Our approach achieves a pass@1 rate of 30.74\%, while ReProver reached 28.28\%, indicating that our method is more effective.
This means that retrieval-augmented language models are effective: For tasks such as formal theorem proving, which relies on external theorem libraries, an effective retrieval model is expected to enhance generation performance.

On test sets with high premise frequency but relatively short proof, our model performs better than ReProver, demonstrating the advantages of a more powerful retriever. In contrast, on test sets with long proof but low premise frequency, our model performs almost on par with ReProver. 
However, on the RD test sets, our model performs slightly worse than ReProver, despite the improvement in retrieval accuracy.
We attribute this outcome to several factors. 
First, incorporating retrieved premises into the prompt increases the length of the context that the model must process compared to providing only the proof state. 
This imposes certain requirements on the model’s parameter size and architecture, as smaller models may struggle to accurately identify useful information within longer contexts. 
We will explore this issue in our future work.

%% file: sections/conclusion.tex
\begin{figure}[h]
    \centering
    \includegraphics[width=0.9\linewidth]{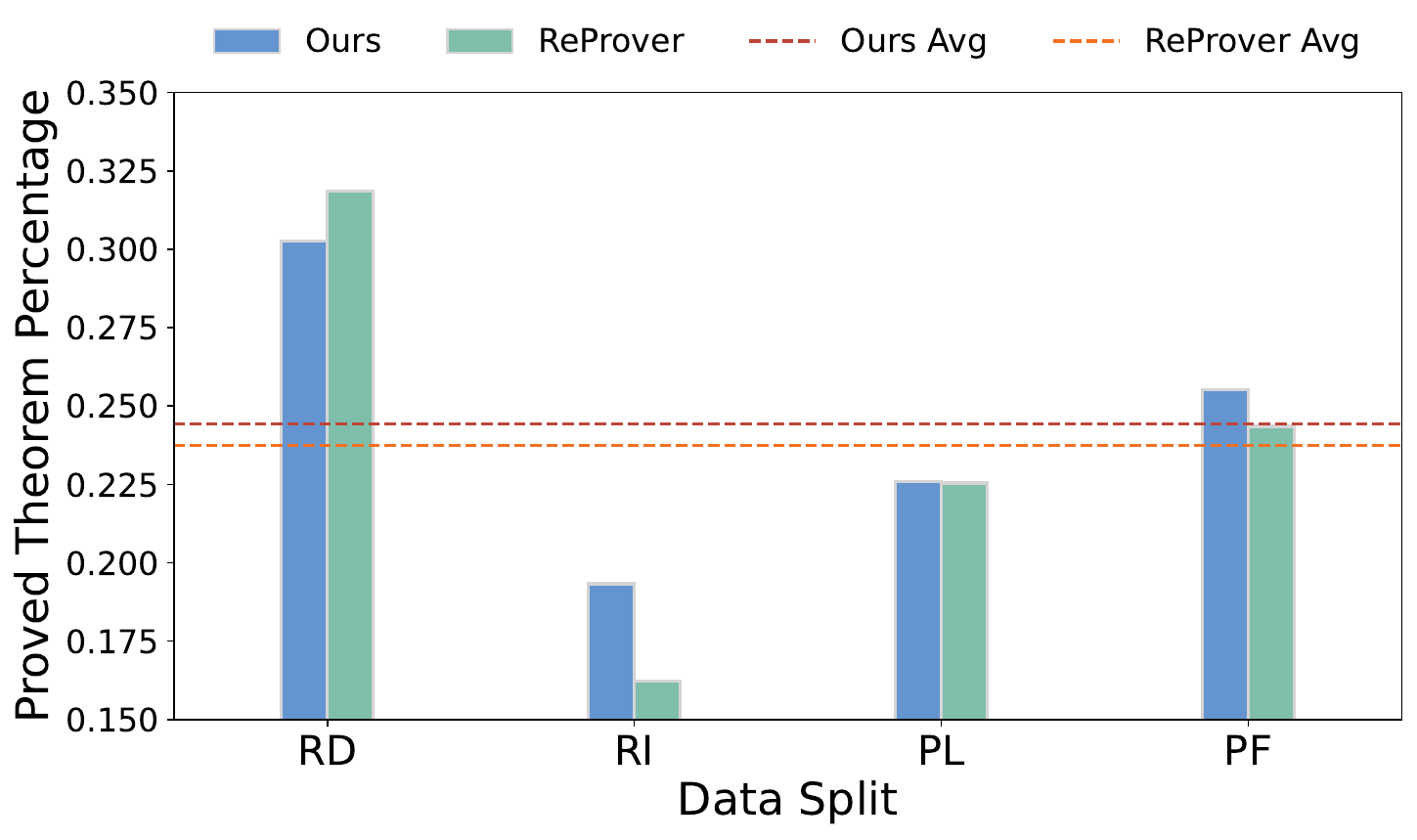}
    \caption{Retrieval-augmented theorem proving results.}
    \label{fig:prover result}
\end{figure}

\section{Conclusion}

In this paper, we propose an innovative method that leverages data extracted from Mathlib4 to train a retrieval model specifically designed for Lean premise retrieval. 
Experiments show that our method outperforms previous models and demonstrates the potential of domain-specific smaller models in data-sparse tasks, such as formal premise retrieval. We have developed a search engine for premise retrieval based on our model. 
We hope that our work can help accelerate mathematical formalization and contributes to the researchers in the community.

Although our method has yielded promising results, there is still room for optimization in the retriever's backbone model.
For the Lean premise retrieval task, the data corpus contains valuable semantic information, such as the hierarchy of Lean’s type system, that has yet to be fully explored. 
Moreover, the results of theorem proving experiments need further clarification with stronger provers, such as DeepSeek-Prover-V2~\cite{ren2025deepseekproverv2advancingformalmathematical}, to demonstrate the effects of premise selection for automated theorem proving.
In the future, we will focus on leveraging this untapped semantic information more effectively and developing better models.


%% file: sections/appendix.tex
\section{Data Extraction}\label{data extraction}
\begin{figure}[h]
    \centering
    \includegraphics[width=\linewidth]{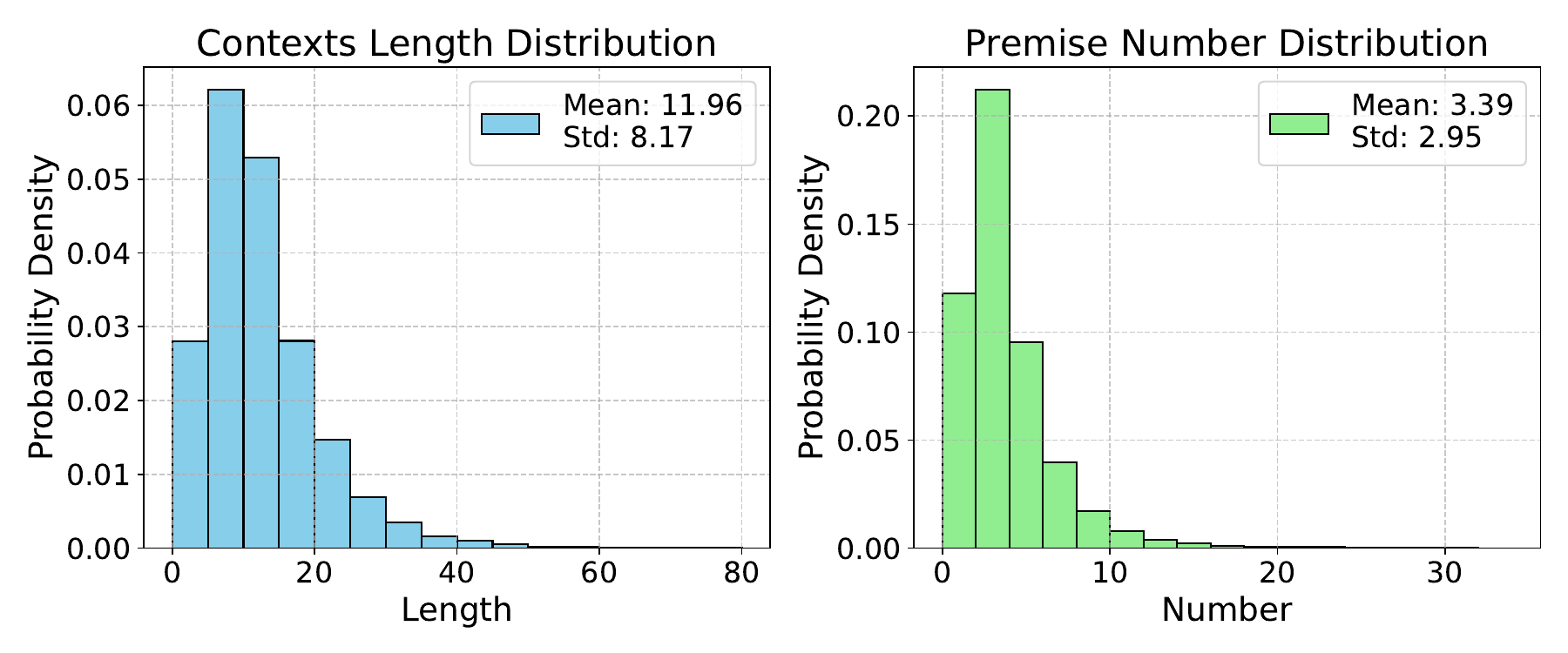}
    \caption{The probability density distribution of context lengths and premise numbers in the extracted dataset.}
    \label{fig:data info}
\end{figure}

Several tools have been developed to extract data from a Lean project. To meet our requirements, we utilize the script from LeanDojo and incorporate additional features. 
We applied this pipeline to Mathlib\footnote{We used tag v4.10.0 of mathlib4. \url{https://github.com/leanprover-community/mathlib4/tree/v4.10.0}}. The extracted information covers all the premises and proofs in this project, along with metadata such as file dependencies. For each premise, we extract its argument list and output. The proofs are lists of tactics, with the state before/after the tactics and premises used in each step. 
Figure~\ref{fig:data info} illustrates the statistics related to context length and the number of premises in the dataset. 
The context length represents the length of $\Gamma$, while the premise number indicates how many premises are utilized in a specific tactic. 
It is important to note that tactics that do not have any premises are excluded from these statistics.

